\documentclass[10pt]{article} 

\usepackage[preprint]{rlj} 

\usepackage{amssymb}            
\usepackage{mathtools}          
\usepackage{mathrsfs}           
\usepackage{graphicx}           
\usepackage{subcaption}         
\usepackage[space]{grffile}     
\usepackage{url}                
\usepackage{lipsum}             
\usepackage{xcolor}

\usepackage{booktabs}

\definecolor{mygreen}{HTML}{00b339}
\definecolor{myred}{HTML}{e41b65}
\definecolor{myorange}{HTML}{f68223}

\definecolor{Background}{RGB}{9,149,255}
\definecolor{StickFigureBackground}{RGB}{209,236,255}
\definecolor{Environment}{RGB}{188,223,255}
\definecolor{MemoryAttention}{RGB}{165,198,243}

\definecolor{InputImages}{RGB}{189,223,229}
\definecolor{TrackingPrompt}{RGB}{208,188,240}
\definecolor{ImageEncoder}{RGB}{255,194,196}
\definecolor{GroundTruth}{RGB}{247,186,208}
\definecolor{MaskDecoder}{RGB}{252,214,182}
\definecolor{MemoryEncoder}{RGB}{241,193,232}
\definecolor{RLPolicy}{RGB}{51,255,116}

\title{SAM2RL: Towards Reinforcement Learning Memory Control in Segment Anything Model 2}

\setrunningtitle{SAM2RL: Towards Reinforcement Learning Memory Control in SAM 2}

\author{Alen Adamyan\textsuperscript{1,2,$\dagger$}, Tomáš Čížek\textsuperscript{1,$\dagger$}, Matej Straka\textsuperscript{1}, Klara Janouskova\textsuperscript{2}, \\Martin Schmid\textsuperscript{1,3}}

\emails{alen.am0161@gmail.com, \ \{cizek,schmidm\}@kam.mff.cuni.cz, matej.straka825@student.cuni.cz, klara.janouskova@fel.cvut.cz}

\affiliations{
$^{1}$\textbf{Department of Applied Mathematics, Charles University, Prague, Czech Republic}\\
$^{2}$\textbf{Visual Recognition Group, Czech Technical University in Prague}\\
$^{3}$\textbf{EquiLibre Technologies, Inc.}\\
\par 
$^\dagger$ Equal contribution.
}

\contribution{
    Provide a succinct but precise list of the contribution(s) of the paper. Use contextual notes to avoid implications of contributions more significant than intended and to clarify and situate the contribution relative to prior work (see the examples below). If there is no additional context, enter ``None''. Try to keep each contribution to a single sentence, although multiple sentences are allowed when necessary. If using complete sentences, include punctuation. If using a single sentence fragment, you may omit the concluding period. A single contribution can be sufficient, and there is no limit on the number of contributions. Submissions will be judged mostly on the contributions claimed on their cover pages and the evidence provided to support them. Major contributions should not be claimed in the main text if they do not appear on the cover page. Overclaiming can lead to a submission being rejected, so it is important to have well-scoped contribution statements on the cover page.
    }
    {
    None
    }

\contribution{
    The submission template for submissions to RLJ/RLC 2025
    }
    {
    Built from previous RLC/RLJ, ICLR, and TMLR submission templates
    }

\contribution{
    \textit{{[}Example of one contribution and corresponding contextual note for the paper ``Policy gradient methods for reinforcement learning with function approximation'' \citep{Sutton2000}.]}\\ This paper presents an expression for the policy gradient when using function approximation to represent the action-value function.
    }
    {
    Prior work established expressions for the policy gradient without function approximation \citep{Williams1992}.
    }


\summary{The summary appears on the cover page. Although it can be identical to the abstract, it does not have to be. One might choose to omit the stated contributions in the Summary, given that they will be stated in the box below. The original abstract may also be extended to two paragraphs. The authors should ensure that the contents of the cover page fit entirely on a single page. The cover page does \textbf{not} count towards the 8--12 page limit.

\lipsum[1]
}

\begin{document}

\maketitle  

\begin{abstract}

Segment Anything Model 2 (SAM~2) has demonstrated strong performance in object segmentation tasks and has become the state-of-the-art for visual object tracking.
The model stores information from previous frames in a memory bank, enabling temporal consistency across video sequences.
Recent methods augment SAM~2 with hand-crafted update rules to better handle distractors, occlusions, and object motion.
We propose a fundamentally different approach using reinforcement learning for optimizing memory updates in SAM~2 by framing memory control as a sequential decision-making problem.
In an overfitting setup with a separate agent per video, our method achieves a relative improvement over SAM~2 that exceeds by more than three times the gains of existing heuristics.
These results reveal the untapped potential of the memory bank and highlight reinforcement learning as a powerful alternative to hand-crafted update rules for memory control in visual object tracking.

\end{abstract}


\section{Introduction}
\label{sec:intro}

\emph{Reinforcement learning} (RL) is well-suited for sequential decision-making tasks in dynamic environments by learning from delayed rewards to optimize long-term returns. 
\emph{Visual Object Tracking} (VOT) poses precisely this challenge, requiring a tracker to localize a target
in a continuous video stream while adapting to appearance changes, occlusions, and distractions over time.
These challenges align naturally with the strengths of RL in learning adaptive policies under uncertainty.

The current state-of-the-art in visual object tracking, \emph{Segment Anything Model 2} (SAM 2) \citep{ravi2024sam2}, integrates a promptable segmentation framework with a memory bank.
The memory bank always retains memories from the most recent frames to maintain temporal consistency.
However, this fixed update rule overlooks each frame’s relevance to the tracking quality.
Building upon SAM~2, several recent studies have proposed heuristic enhancements to this memory bank update mechanism, leading to notable performance improvements.
These heuristics target specific scenarios where SAM~2 underperforms, such as motion-aware selection \citep{yang2024samurai, yang2025mosam} and distractor-aware updates \citep{videnovic2024distractor}.

Rather than hand-crafting these update rules, we propose to learn them directly by framing the SAM~2 memory updates as a reinforcement learning task.
We estimate the upper-bound performance achievable with learned memory updates by overfitting an RL agent (SAM2RL) to a set of videos individually.
In this overfitting setup, SAM2RL achieves more than three times the performance gains of prior methods over the original SAM~2, highlighting the potential of RL to improve SAM~2 memory bank updates.
This improvement could set a new state-of-the-art for VOT without modifying any trainable parameters of the original SAM~2.

The paper is organized as follows.
In Section \ref{sec:related}, we define the visual tracking task and review relevant prior work.
In Section \ref{sec:method}, we describe the SAM~2 architecture and formulate memory bank updates as a reinforcement learning task.
In Section \ref{sec:evaluation}, we overfit our policy on a set of videos and compare its performance against prior methods.
Finally, Section \ref{sec:conclusion} concludes with a discussion of our findings and outlines directions for future work.

\section{Related Work}
\label{sec:related}

\emph{Vision Object Tracking} (VOT) is the task of locating and following a specific object over time in a video sequence, given its initial position in the first frame \citep{tracking_dl_survey}.
Traditionally, this task has been approached using bounding boxes to represent the target object, offering a simple yet effective way to estimate object location and scale.
In recent years, the field has shifted toward more precise localization methods, with tracking-by-segmentation becoming the dominant approach.
Common challenges of VOT include occlusion, motion blur, and appearance changes.

\emph{Segment Anything Model 2} (SAM 2) \citep{ravi2024sam2} extends the promptable transformer architecture of \emph{Segment Anything Model} (SAM) \citep{kirillov2023segment} with a streaming memory mechanism, enabling unified image and video segmentation.
Leveraging large-scale pretraining, it enables real-time zero-shot tracking across diverse domains, such as self-driving, video editing, or wildlife monitoring.
SAM 2's fixed-length memory bank always stores only the most recent frames, suffering from redundancy and irrelevant content.
To address these limitations, several recent extensions propose improved memory update mechanisms using hand-crafted update rules.

\emph{SAM2Long} \citep{ding2024sam2long} mitigates error accumulation in long videos by using multiple parallel memory banks, improving occlusion handling and object reappearance at increased computational cost.
\emph{DAM4SAM} \citep{videnovic2024distractor} enhances distractor handling with distractor-aware update rules.
\emph{SAMURAI} \citep{yang2024samurai} incorporates motion-aware memory selection via a \emph{Kalman filter} \citep{kalman1960new} for refined mask selection, while \emph{MoSAM} \citep{yang2025mosam} introduces motion-guided prompting and enhances the memory bank with spatial-temporal memory selection for accurate predictions.

\section{Method}
\label{sec:method}
In this section, we formulate memory selection in SAM~2 as a reinforcement learning task, where an agent learns a memory update policy to optimize tracking performance.
We first give a brief overview of the SAM~2 architecture, and then describe how memory selection is framed as a reinforcement learning task.

\begin{figure}[!htb]
    \centering
\includegraphics[width=1.0\linewidth]{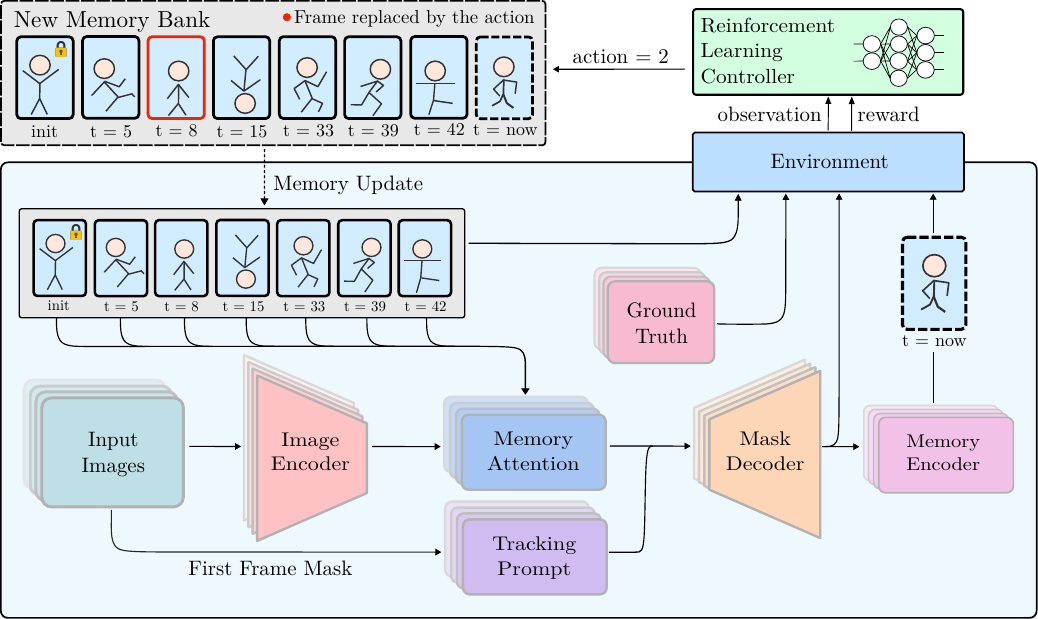}
    \caption{
    In SAM2RL, the image encoder processes a new frame, which is then conditioned by the
    memory bank of past predictions and tracking prompts. 
    The mask decoder uses this representation to generate a segmentation mask.
    This mask is used to provide a reward to the RL memory controller and also by the memory encoder to create a new memory for the current frame.
    The controller then observes memories and decides which memory frames to retain.
    }
    \label{fig:sam2rl}
\end{figure}

\paragraph{SAM 2 Architecture.}
SAM~2 achieves temporally consistent segmentation of a target object by combining per-frame image embeddings with a memory-attention mechanism.
As illustrated in Figure \ref{fig:sam2rl}, each incoming frame is first processed by the \emph{image encoder} to produce a feature embedding.
This embedding is then conditioned on \emph{memories} inside the \emph{memory bank} using the \emph{memory attention} and also on \emph{tracking prompts}, such as the input mask of the first frame.
The resulting conditioned embedding is passed into the \emph{mask decoder}, which generates the final segmentation mask.
The \emph{memory encoder} then transforms the predicted mask and the corresponding image embedding into a \emph{memory feature}.
This memory feature is added to the memory bank with fixed capacity $N = 7$, which always retains the initial frame along with the most recent $N-1$ memories.

\paragraph{Memory Selection via Reinforcement Learning.}
We frame the SAM~2 memory selection problem as a reinforcement learning task.
The tracking of a single video object forms an episode in the RL environment operated by an agent who performs memory bank updates.
At each timestep, the incoming input image is first processed by the frozen SAM~2 pipeline, producing the predicted mask and the encoded memory feature.
The agent then receives \emph{observation} consisting of the representation of the stored memories in the memory bank and the memory feature of the current frame.
Based on this observation, the agent selects an \emph{action} that either discards the incoming memory feature or inserts it into the memory bank by replacing one of the $N-1$ stored memories.
The objective is to learn a memory update policy that optimizes the \emph{tracking quality} $Q$, which is the primary tracking measure of the VOTS challenge \citep{vots}.
The tracking quality is measured as the mean \emph{intersection-over-union}~(IoU) on frames with a visible object, with a penalty for incorrect prediction when the object is not present.
Formally, the tracking quality $Q$ for a video of length $T$ with a predicted mask $\hat{M}_{t}$ and a ground truth mask $M_{t}$ at timestep $t$ is defined as
\begin{equation}
\label{eq:tracking_quality}
Q = \frac{1}{T} \sum_{t=1}^{T} q_{t}, \quad \text{where} \quad
q_{t} =
\begin{cases}
1, & \hat{M}_{t} = \emptyset \text{ and } M_{t} = \emptyset, \\[3pt]
\mathrm{IoU}(\hat{M}_{t}, M_{t}), & \text{otherwise}.
\end{cases}
\end{equation}
To guide the learning process towards maximizing the tracking quality $Q$, we set the \emph{reward} at timestep $t$ to~$q_{t}$.
The final episode \emph{return} to maximize is thus equal to $\sum_{t=1}^{T}{\gamma^{t-1}q_t}$ with $\gamma \in (0, 1]$.

\section{Evaluation}
\label{sec:evaluation}

The memory bank is a relatively small component of the overall SAM~2 architecture.
Therefore, it remains unclear how much of the tracking performance, even on the training dataset, can be improved by optimizing only the memory update policy.
In this section, we explore the power of the memory bank and its update policy by overfitting a sequence of videos individually using a reinforcement learning algorithm, as described in Section~\ref{sec:method}.
We compare the obtained results with those achieved by previous methods using hand-crafted update rules to show how much of the capacity is effectively utilized by the current methods.

\begin{table}[!htb]
    \centering
    \caption{Comparison of SAM~2-based methods and the proposed SAM2RL with an overfitted memory bank policy, evaluated on 64 videos from the SA-V training set using standard VOTS metrics.}
    \label{tab:metricsEvaluation}
    \begin{tabular}{lrrr}
        \toprule
        \textbf{Method} & \multicolumn{1}{c}{\textbf{Quality [\%]}} & \multicolumn{1}{c}{\textbf{Accuracy [\%]}} & \multicolumn{1}{c}{\textbf{Robustness [\%]}} \\
        \midrule
        SAM~2   & 71.95 & 73.01 & 88.84 \\        
        \midrule
        SAMURAI  & {\scriptsize\textcolor{myred}{-4.70}} 67.25 & {\scriptsize\textcolor{myred}{-0.30}} 72.71 & {\scriptsize\textcolor{mygreen}{+1.83}} 90.67 \\
        DAM4SAM  & {\scriptsize\textcolor{mygreen}{+1.60}} 73.55 & {\scriptsize\textcolor{myred}{-0.05}} 72.96 & {\scriptsize\textcolor{mygreen}{+0.36}} 89.20 \\
        SAM2RL (ours) & \textbf{{\scriptsize\textcolor{mygreen}{+4.91}} 76.86} & \textbf{{\scriptsize\textcolor{mygreen}{+2.52}} 75.53} & \textbf{{\scriptsize\textcolor{mygreen}{+2.33}} 91.17} \\
        \bottomrule
    \end{tabular}
\end{table}

\begin{figure}[!htb]
    \centering
    \includegraphics[scale=1]{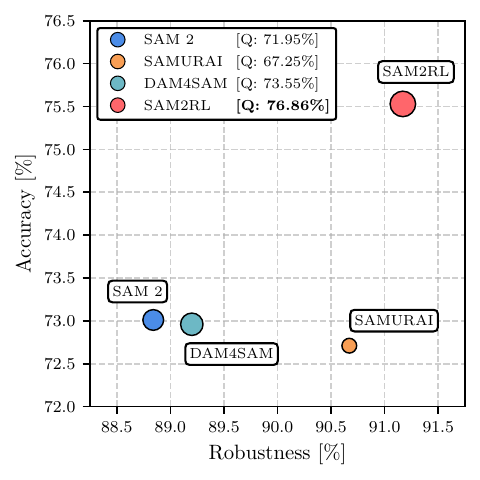}
    \caption{Accuracy vs. robustness plot comparing the performance of SAM~2-based models and SAM2RL, with marker size indicating the tracking quality.}
    \label{fig:ar_plot}
\end{figure}

\begin{figure}[!htb]
    \centering
    \includegraphics[scale=0.93]{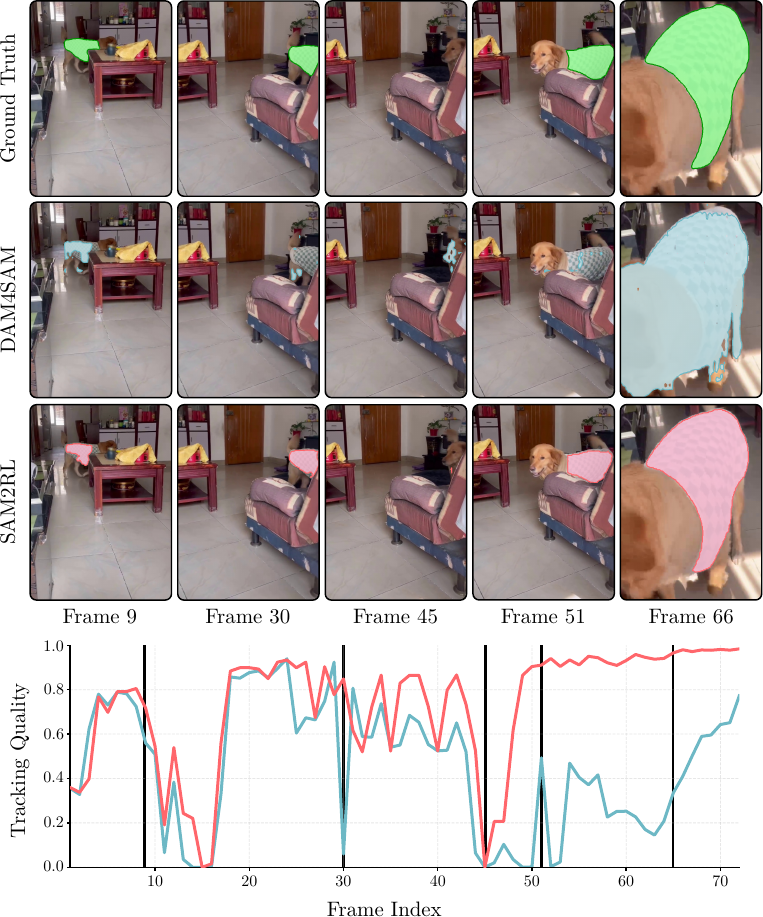}
    \caption{Visualization of predicted masks and the tracking quality of SAM2RL and DAM4SAM on a single selected video.
    SAM2RL produces masks that more closely match the ground truth, and successfully recovers the object after its disappearance, while DAM4SAM often fails to detect it.
    }
    \label{fig:dog_plot}
\end{figure}

\paragraph{Training Setup.}
Our experiments are designed to intentionally overfit individual videos, without any expectation that the learned policy will generalize beyond them.
To this end, we represent the agent's observation using a simple binary vector that uniquely encodes both the indices of the image frames stored in the memory bank and the current timestep.
The policy network is then a two-layer multilayer perceptron (MLP), with an input size equal to the video length, a hidden layer of 1024 units, and a softmax output over the $N$ possible actions.
We train the network using \emph{Proximal Policy Optimization} (PPO) \citep{ppo} separately on 64 random videos from the \emph{SA-V training dataset} \citep{ravi2024sam2}, with an average length of 80 frames recorded at 6 frames per second.
All evaluated models employ the largest available \emph{Hiera-L} image encoder \citep{hiera} with input resolution $256 \times 256$ and SAM~2.1 checkpoint.
We always track only the first object in a video.

We evaluate the environment and train the policy network for each video using a single NVIDIA A100 40GB GPU.
The environment simulation runs at approximately 750 frames per second.
In each PPO iteration, we collect a dataset of 16k samples and train the policy for 2 epochs.
The training process runs for 150 iterations and takes approximately 1 hour on average per video.

\paragraph{Quantitative Results.}
We compare the tracking performance of the proposed SAM2RL with SAM~2 and its two best-performing publicly available variants, SAMURAI \citep{yang2024samurai} and DAM4SAM \citep{videnovic2024distractor}.
We evaluate the tracking performance in Table~\ref{tab:metricsEvaluation} using three standard metrics from the VOTS challenge \citep{vots}.
The tracking quality $Q$ from Equation~\eqref{eq:tracking_quality} is the primary performance measure optimized by the RL policy.
The \emph{accuracy} and \emph{robustness} are two additional metrics to evaluate mask quality on frames with a visible object only.
The accuracy is equal to the mean IoU in frames with nonzero IoU, and the robustness is the proportion of frames with nonzero IoU in frames with a visible object.
The accuracy and robustness thus describe the trade-off between the mask precision and the chance of detecting an object with at least a minimal overlap.
We visualize all the metric values in Figure~\ref{fig:ar_plot}.

SAM2RL achieves the highest performance in all three VOTS metrics, surpassing SAM~2 and both of its enhanced variants.
In particular, SAM2RL improves the tracking quality by +4.91\%, accuracy by +2.52\%, and robustness by +2.33\% compared to SAM~2.
In contrast, SAMURAI, in our training setup, underperforms relative to SAM~2 in tracking quality metric by -4.70\%, which we attribute to its tendency to predict a mask even when the object is not visible, trading off for a modest gain in robustness by +1.83\%.
DAM4SAM shows a moderate improvement over SAM~2 with a +1.60\% increase in tracking quality, but it still falls short compared to SAM2RL.
These results indicate that optimizing the memory bank's update policy can lead to substantial improvements in tracking performance, exceeding those achieved by current methods.

\paragraph{Qualitative Results.}
Figure~\ref{fig:dog_plot} presents a qualitative comparison of the predicted masks generated by SAM2RL and DAM4SAM in a selected video from the SA-V dataset.
DAM4SAM is used as a baseline, as it outperforms both SAM~2 and SAMURAI on this specific video.
SAM2RL produces masks that match the ground truth more accurately than those from DAM4SAM.
In particular, after the tracked object disappears completely in frame 45, SAM2RL successfully recovers object tracking, whereas DAM4SAM either fails to detect the object completely or produces inaccurate masks.
On the other hand, we observe that although SAM2RL is deliberately overfitted on this video, it still fails to achieve optimal performance, suggesting the limit of the memory bank capacity.

\section{Conclusion}
\label{sec:conclusion}

We introduced the first reinforcement learning approach to optimize memory update policies within SAM~2 for visual object tracking.
Although the memory bank constitutes only a small component of the overall SAM~2 architecture, the significant performance gains of SAM2RL in an overfitting setup demonstrate that optimizing the update policy can greatly enhance the tracking quality.
This highlights an opportunity to improve the current hand-crafted update rules using reinforcement learning-based update policies.
We plan to achieve these improvements by scaling our approach through the integration of richer observations and suitable network architectures, enabling us to learn policies that would generalize beyond the training data, and thus push the current state-of-the-art in video object tracking.

\subsubsection*{Acknowledgments}
\label{sec:ack}
Tomáš Čížek and Martin Schmid were supported by the grant no. 25-18031S of the Czech Science Foundation (GAČR).
Martin Schmid was also supported by Center for Foundations of Modern Computer Science
(Charles Univ. project UNCE 24/SCI/008).
Klara Janouskova received support from Toyota Motor Europe, and Alen Adamyan from CEDMO 2.0 NPO (MPO 60273/24/21300/21000).
Computational resources were provided by the e-INFRA CZ project (ID:90254), supported by the Ministry of Education, Youth and Sports of the Czech Republic.



\bibliography{main}

\begin{thebibliography}{11}
\providecommand{\natexlab}[1]{#1}
\providecommand{\url}[1]{\texttt{#1}}
\expandafter\ifx\csname urlstyle\endcsname\relax
  \providecommand{\doi}[1]{DOI: #1}\else
  \providecommand{\doi}{DOI: \begingroup \urlstyle{rm}\Url}\fi

\bibitem[Ding et~al.(2024)Ding, Qian, Dong, Zhang, Zang, Cao, Guo, Lin, and Wang]{ding2024sam2long}
Shuangrui Ding, Rui Qian, Xiaoyi Dong, Pan Zhang, Yuhang Zang, Yuhang Cao, Yuwei Guo, Dahua Lin, and Jiaqi Wang.
\newblock {S}am2{L}ong: {E}nhancing {S}am 2 for {L}ong {V}ideo {S}egmentation with a {T}raining-{F}ree {M}emory {T}ree.
\newblock \emph{arXiv preprint arXiv:2410.16268}, 2024.

\bibitem[Kalman(1960)]{kalman1960new}
Rudolph~Emil Kalman.
\newblock A {N}ew {A}pproach to {L}inear {F}iltering and {P}rediction {P}roblems, 1960.

\bibitem[Kirillov et~al.(2023)Kirillov, Mintun, Ravi, Mao, Rolland, Gustafson, Xiao, Whitehead, Berg, Lo, et~al.]{kirillov2023segment}
Alexander Kirillov, Eric Mintun, Nikhila Ravi, Hanzi Mao, Chloe Rolland, Laura Gustafson, Tete Xiao, Spencer Whitehead, Alexander~C Berg, Wan-Yen Lo, et~al.
\newblock Segment {A}nything.
\newblock In \emph{Proceedings of the {IEEE/CVF} {I}nternational {C}onference on {C}omputer {V}ision}, pp.\  4015--4026, 2023.

\bibitem[Kristan et~al.(2023)Kristan, Matas, Danelljan, Felsberg, Chang, Zajc, Luke{\v{z}}i{\v{c}}, Drbohlav, Zhang, Tran, et~al.]{vots}
Matej Kristan, Ji{\v{r}}{\'\i} Matas, Martin Danelljan, Michael Felsberg, Hyung~Jin Chang, Luka~{\v{C}}ehovin Zajc, Alan Luke{\v{z}}i{\v{c}}, Ondrej Drbohlav, Zhongqun Zhang, Khanh-Tung Tran, et~al.
\newblock The {F}irst {V}isual {O}bject {T}racking {S}egmentation {VOTS2023} {C}hallenge {R}esults.
\newblock In \emph{Proceedings of the IEEE/CVF International Conference on Computer Vision}, pp.\  1796--1818, 2023.

\bibitem[Marvasti-Zadeh et~al.(2022)Marvasti-Zadeh, Cheng, Ghanei-Yakhdan, and Kasaei]{tracking_dl_survey}
Seyed~Mojtaba Marvasti-Zadeh, Li~Cheng, Hossein Ghanei-Yakhdan, and Shohreh Kasaei.
\newblock Deep {L}earning for {V}isual {T}racking: A {C}omprehensive {S}urvey.
\newblock \emph{IEEE {T}ransactions on {I}ntelligent {T}ransportation {S}ystems}, 23\penalty0 (5):\penalty0 3943--3968, 2022.
\newblock \doi{10.1109/TITS.2020.3046478}.

\bibitem[Ravi et~al.(2024)Ravi, Gabeur, Hu, Hu, Ryali, Ma, Khedr, R{\"a}dle, Rolland, Gustafson, et~al.]{ravi2024sam2}
Nikhila Ravi, Valentin Gabeur, Yuan-Ting Hu, Ronghang Hu, Chaitanya Ryali, Tengyu Ma, Haitham Khedr, Roman R{\"a}dle, Chloe Rolland, Laura Gustafson, et~al.
\newblock {S}am 2: {S}egment {A}nything in {I}mages and {V}ideos.
\newblock \emph{arXiv preprint arXiv:2408.00714}, 2024.

\bibitem[Ryali et~al.(2023)Ryali, Hu, Bolya, Wei, Fan, Huang, Aggarwal, Chowdhury, Poursaeed, Hoffman, Malik, Li, and Feichtenhofer]{hiera}
Chaitanya Ryali, Yuan-Ting Hu, Daniel Bolya, Chen Wei, Haoqi Fan, Po-Yao Huang, Vaibhav Aggarwal, Arkabandhu Chowdhury, Omid Poursaeed, Judy Hoffman, Jitendra Malik, Yanghao Li, and Christoph Feichtenhofer.
\newblock Hiera: {A} {H}ierarchical {V}ision {T}ransformer without the {B}ells-and-{W}histles.
\newblock \emph{arXiv preprint arXiv:2306.00989}, 2023.

\bibitem[Schulman et~al.(2017)Schulman, Wolski, Dhariwal, Radford, and Klimov]{ppo}
John Schulman, Filip Wolski, Prafulla Dhariwal, Alec Radford, and Oleg Klimov.
\newblock Proximal {P}olicy {O}ptimization {A}lgorithms.
\newblock \emph{arXiv preprint arXiv:1707.06347}, 2017.

\bibitem[Videnovic et~al.(2024)Videnovic, Lukezic, and Kristan]{videnovic2024distractor}
Jovana Videnovic, Alan Lukezic, and Matej Kristan.
\newblock A {D}istractor-{A}ware {M}emory for {V}isual {O}bject {T}racking with {SAM2}.
\newblock \emph{arXiv preprint arXiv:2411.17576}, 2024.

\bibitem[Yang et~al.(2024)Yang, Huang, Chai, Jiang, and Hwang]{yang2024samurai}
Cheng-Yen Yang, Hsiang-Wei Huang, Wenhao Chai, Zhongyu Jiang, and Jenq-Neng Hwang.
\newblock {S}amurai: {A}dapting {S}egment {A}nything {M}odel for {Z}ero-{S}hot {V}isual {T}racking with {M}otion-{A}ware {M}emory.
\newblock \emph{arXiv preprint arXiv:2411.11922}, 2024.

\bibitem[Yang et~al.(2025)Yang, Yao, Cui, and Bo]{yang2025mosam}
Qiushi Yang, Yuan Yao, Miaomiao Cui, and Liefeng Bo.
\newblock {M}o{SAM}: {M}otion-{G}uided {S}egment {A}nything {M}odel with {S}patial-{T}emporal {M}emory {S}election.
\newblock \emph{arXiv preprint arXiv:2505.00739}, 2025.

\end{thebibliography}
\bibliographystyle{rlj}



\end{document}